\documentclass[letterpaper]{article} 
\usepackage[preprint]{aaai2027}  
\usepackage[hyphens]{url}  
\usepackage{graphicx} 
\usepackage{natbib}  
\usepackage{caption} 
\usepackage{algorithm}
\usepackage{algorithmic}

\usepackage{newfloat}
\usepackage{listings}
\DeclareCaptionStyle{ruled}{labelfont=normalfont,labelsep=colon,strut=off} 
\floatstyle{ruled}
\newfloat{listing}{tb}{lst}{}
\floatname{listing}{Listing}

\usepackage{booktabs,multirow}

\usepackage{amsmath,amssymb,bm}

\title{Competition Domains Matter: Selective Head Globalization for Training-Free Sparse Attention in Video Diffusion Transformers}
\author{Tiantian Liu, Meng Wan, Zhiqiang Liang, Jue Wang$^{\ast}$, Ningming Nie, Yangang Wang$^{\ast}$}

\affiliations{University of Chinese Academy of Sciences, Beijing, China \\ $^{\ast}$Corresponding authors.}

\begin{document}

\setcitestyle{authoryear,aysep={ },yysep={; },notesep={, }}
\let\cite\citep

\maketitle

\begin{abstract}
Training-free sparse attention is a practical route to efficient video diffusion transformer inference, yet existing methods typically fix the competition domain at each query row for every attention head. Prior work primarily addresses how to estimate token-level importance within this fixed domain; whether the domain itself should adapt to head heterogeneity remains under-specified. We introduce \textbf{EG-Sparse}, a training-free framework that routes low-entropy heads to a grouped cross-head pool while retaining per-row local selection for high-entropy heads, using only scores already materialized by the sparse pipeline. A controlled diagnostic fixes the ranking score and varies only the competition rule: grouped cross-head selection reduces low-entropy-head reconstruction error by $9.8\%$, whereas forcing high-entropy heads into an all-head global pool increases it by $7.9\%$. Across four configurations, EG-Sparse achieves higher video quality while reducing attention density and GPU denoising latency compared with strong training-free sparse-attention baselines. Mechanism ablations further show that selective domain adaptation outperforms uniform, random, and fixed domain-assignment policies. These results support competition-domain adaptation as a distinct design axis within centroid-guided training-free sparse attention.
\end{abstract}


\section{Introduction}

Video diffusion transformers (DiTs)~\cite{peebles2023scalable} have become a central backbone of modern video generation, powering state-of-the-art models such as CogVideoX~\cite{yang2025cogvideox}, HunyuanVideo~\cite{kong2025hunyuanvideo}, and Wan~\cite{wan2025wan}.
Recent systems further extend this paradigm toward scalable and long-form video synthesis~\cite{walt2024,pyramidflow2025,stiv2025,streamingt2v2025}.
As these systems scale toward longer videos and higher resolutions, the quadratic complexity of full 3D attention becomes the dominant inference bottleneck.

Efficiency research around diffusion transformers spans architecture/token-level compression~\cite{asymrnr,udits,ptt2iv,sparsedit}, inter-step reuse~\cite{learningtocache,pab,teacache,fastercache}, quantization and distributed execution~\cite{svdquant,distrifusion}, attention-kernel optimization~\cite{flashattention3,sageattention,sageattention2,flashmask}, and head compression~\cite{ditfastattnv2,vsa}.
Our focus is distinct: the competition domain used to select interactions within a centroid-guided sparse-attention pipeline.

Training-free sparse attention has emerged as a practical remedy: it selectively computes only a subset of query-key interactions at inference time without modifying pretrained weights~\cite{svgen,svg2_2025,spargeattn2025,xattention2025}.

Within training-free sparse attention, top-$p$ block selection has become the dominant paradigm, adopted by SVG2~\cite{svg2_2025}, XAttention~\cite{xattention2025}, SpargeAttn~\cite{spargeattn2025}, and others.
The common workflow first partitions tokens into clusters, then ranks candidate (query-cluster, key-cluster) pairs by some importance score, and retains pairs until their cumulative importance reaches a threshold $p$.
Despite extensive work on improving the ranking (how to estimate block importance more accurately), the partition and retention steps share an unexamined assumption: the competition domain, i.e., the set of candidate pairs that compete for the same budget, is always each head independently and within each head, each query row independently. 

\begin{figure}[htbp]
\centering
\includegraphics[width=\columnwidth]{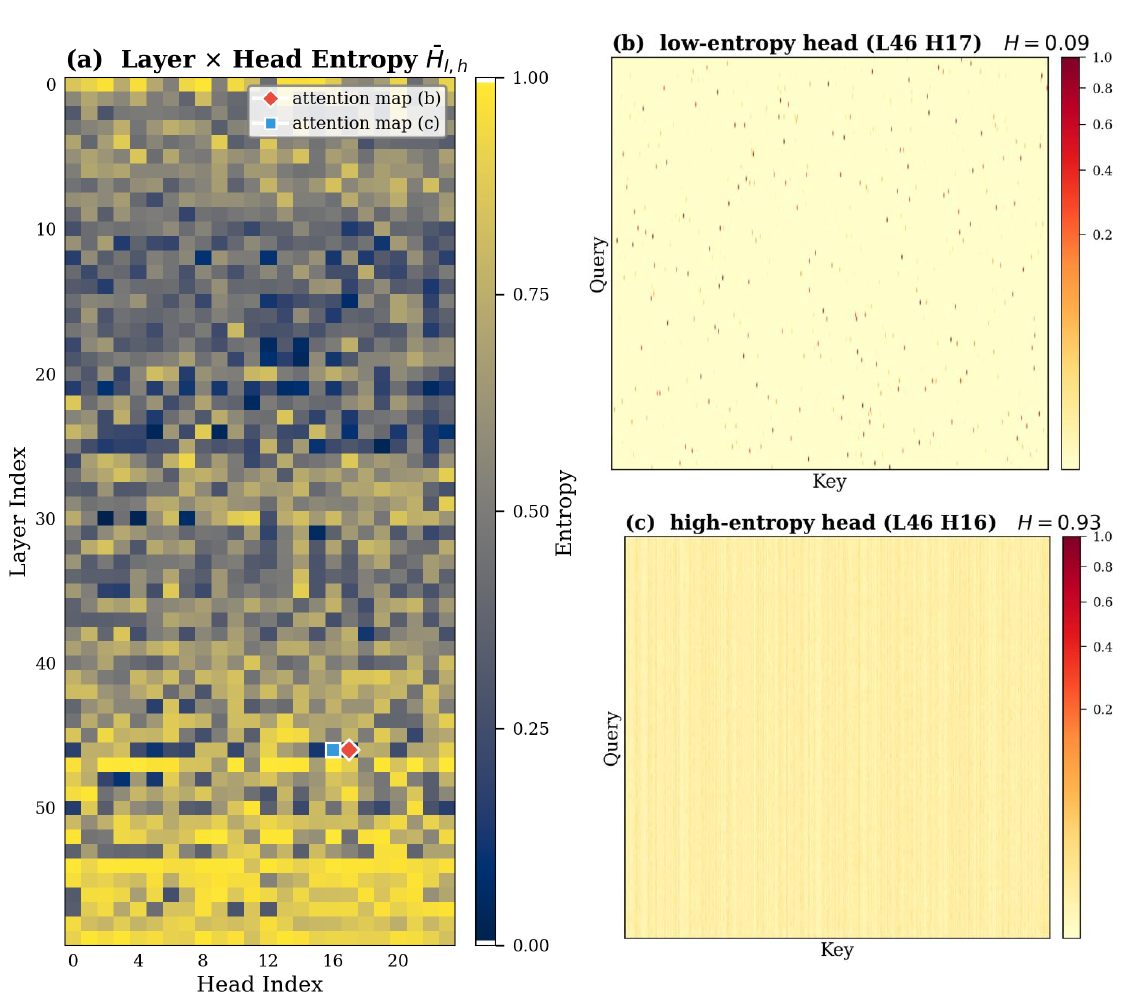}
\caption{\textbf{Attention heads exhibit substantial structural heterogeneity.} (a) Normalized attention entropy $\widetilde H_h=H_h/\log C_k$ varies across layers and heads. (b--c) Two representative heads from the same layer show distinct dense centroid-attention patterns: a low-entropy head concentrates its mass on a small set of query--key cluster pairs, whereas a high-entropy head distributes attention broadly across keys. This heterogeneity motivates head-adaptive competition domains.}
\label{fig:motivation}
\end{figure}

This uniform policy ignores substantial head heterogeneity in video DiTs.
Some heads concentrate most probability mass on a few key clusters, whereas others distribute it broadly.
Figure~\ref{fig:motivation} shows this variation across layers and heads and contrasts two dense attention structures from the same layer.
Normalized entropy varies substantially at both scales: one representative head concentrates on sparse cluster pairs, whereas the other spreads mass broadly across keys.
The value of enlarging the competition domain should therefore depend on head structure.

A controlled diagnostic (Section~3.3) fixes the ranking score and varies only the domain. Low-entropy heads benefit from larger pools, whereas high-entropy heads are penalized in an all-head global pool where concentrated heads capture disproportionate budget. Fixed-domain policies therefore cannot satisfy both groups simultaneously.

\begin{figure*}[t]
\centering
\includegraphics[width=\textwidth]{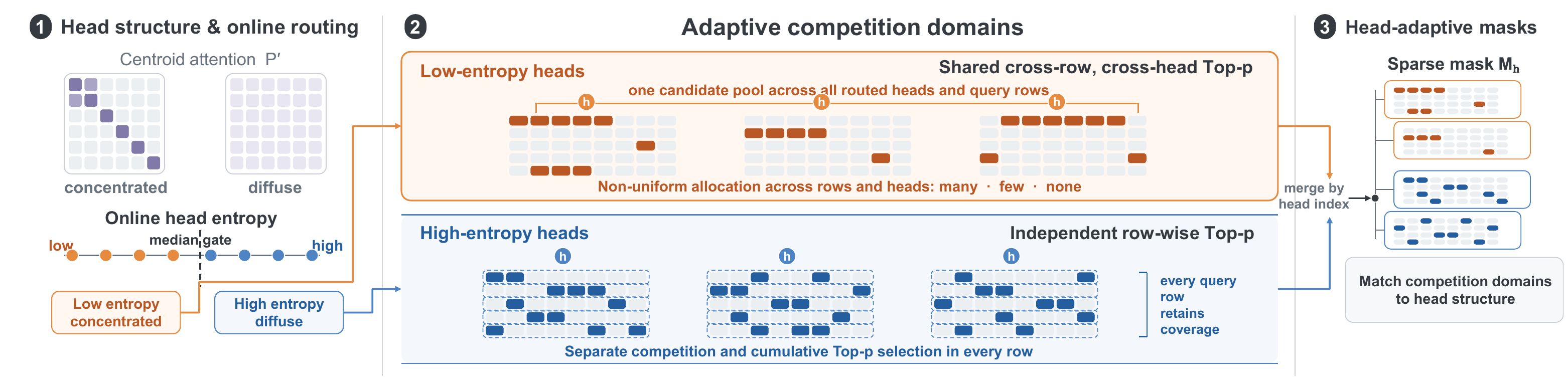}
\caption{\textbf{Overview of EG-Sparse.} A fixed row-wise domain constrains low-entropy heads because the budget cannot flow across rows to where it is most needed, whereas indiscriminate all-head global competition penalizes high-entropy heads. EG-Sparse resolves both: entropy routes low-entropy heads to a grouped cross-head pool, retains per-row local selection for the remaining heads, and merges both masks for execution by the sparse attention kernel.}
\label{fig:overview}
\end{figure*}

To turn this analysis into online competition-domain adaptation, we propose \textbf{EG-Sparse} (Entropy-Guided Sparse Attention). As shown in figure~\ref{fig:overview}, at each denoising step, it computes per-head entropy from centroid attention already materialized by the base pipeline.
Low-entropy heads enter a grouped cross-head pool while the remainder retain per-row local selection.
The current layer's median entropy provides an adaptive threshold without manual tuning, offline calibration, or prompt pools.
EG-Sparse avoids recomputing attention scores: entropy uses $\mathbf{P}'$ already materialized by the base pipeline, and the median split is $O(H \log H)$ with $H = 24$.
The implementation is training-free and operates directly on the centroid-level attention scores already produced by the SVG2 pipeline.

We evaluate EG-Sparse across four model configurations spanning Wan2.1 and HunyuanVideo, covering both text-to-video (T2V) and image-to-video (I2V) generation, at 480P and 720P resolutions.
Across all four configurations, EG-Sparse improves PSNR, SSIM and VBench scores while reducing LPIPS, attention density, and GPU denoising latency.
Controlled ablations separate the contribution of entropy-guided domain assignment from the additional gain of value-aware ranking, showing that the diagnosed mismatch translates into practical improvements without retraining or calibration.

Our contributions are as follows:
\begin{enumerate}
    \item We identify the \textbf{competition-domain mismatch}: a controlled diagnostic at fixed ranking score reveals that head heterogeneity determines opposing domain needs, where low-entropy heads benefit from cross-head budget sharing while high-entropy heads suffer interference when competing in the same pool.
    \item We propose \textbf{EG-Sparse}, a training-free online framework that selectively expands the competition domain to cross-head global selection for low-entropy heads while retaining per-row local selection for high-entropy heads.
    \item We demonstrate consistent \textbf{quality--efficiency improvements} over strong baselines across four model configurations. Mechanism ablations validate that entropy-guided selective domain adaptation outperforms uniform, random and fixed domain-assignment policies.
\end{enumerate}

\section{Related Work}

\paragraph{Video-DiT sparse attention.}
Sparse VideoGen (SVG) profiles spatio-temporal head patterns~\cite{svgen}; SVG2 adds semantic token clustering and centroid Top-$p$ selection~\cite{svg2_2025}; XAttention uses anti-diagonal block scores~\cite{xattention2025}; and SpargeAttn applies selective token compression~\cite{spargeattn2025}.
Sparse-vDiT and Compact Attention search offline for structured, model-specific sparse patterns~\cite{sparsevdit,compactattention}, whereas EG-Sparse routes heads online from current-layer entropy without offline profiling.
Other video methods change sliding-tile locality or cached search~\cite{sta,adaspa}, structured topology or token order~\cite{radialattention,liteattention,paroattention}, and low-resolution guidance or progressive resolution~\cite{draftattention,jenga}.
These approaches primarily change how candidates are found or executed; EG-Sparse changes which candidates share a Top-$p$ budget.



\paragraph{Head adaptation and cross-domain context.}
Video-DiT methods allocate head-wise computation using training statistics, learned mechanisms, or offline calibration~\cite{dsv,ditfastattnv2,haste}.
AdaSpa combines cached search with head-adaptive sparsity~\cite{adaspa}, whereas HASTE couples temporal reuse with calibrated per-head budgets.
EG-Sparse instead uses current-step entropy to decide which heads share a pool, without learned routing or calibration.
Entropy has previously been diagnostic~\cite{rtpurbo,usv}; here it predicts domain-dependent efficiency.
Head-role and adaptive-selection methods for causal language models~\cite{duoattention,seerattention,twilight2025,tactic2025} assume different masks and inference dynamics, so they provide conceptual context rather than direct video-DiT baselines.

\section{Method}

\subsection{Preliminaries}

We follow the standard centroid-guided setup of modern video sparse attention~\cite{svg2_2025,spargeattn2025}.
At a given layer and denoising step, the $H$ attention heads process query and key tokens that are clustered (for instance, by $k$-means) into $C_q$ query centroids and $C_k$ key centroids.
Let $n_{i}$ and $n_{j}$ denote the number of tokens in query cluster $i$ and key cluster $j$, respectively.
For head $h$, the centroid-level attention is computed via the cluster-size-weighted softmax:

\begin{equation}
    P'_{h,i,j}= \frac{n_{j}\,\exp\!\big(S_{h,i,j}\big)}
    {\sum_{k=1}^{C_k} n_{k}\,\exp\!\big(S_{h,i,k}\big)},
    \qquad
    S_{h,i,j}= \frac{\bm{q}_{h,i}^{\top}\bm{k}_{h,j}}{\sqrt{d}} ,
    \label{eq:p_prime}
\end{equation}
where $\bm{q}_{h,i}$ and $\bm{k}_{h,j}$ are the centroids.
Each row sums to one, $\sum_{j}P'_{h,i,j}=1$; $P'_{h,i,j}$ approximates the total attention mass that a representative query in cluster $i$ allocates to the entire key cluster $j$.
Sparse attention selects a subset of cluster pairs to compute densely and skips the rest.

\paragraph{The head-wise paradigm.}
Given a per-token target $p$, existing methods select, \emph{independently for each} $(h,i)$, the smallest set of key clusters whose cumulative mass reaches $p$:

\begin{equation}
    \mathcal{S}_{h,i}=\operatorname*{Top}_{p}\big(P'_{h,i,:}\big),\qquad
    \text{s.t.}\;\; \sum_{j\in\mathcal{S}_{h,i}} P'_{h,i,j}\ge p .
    \label{eq:headwise}
\end{equation}

Equation~\eqref{eq:headwise} is the \textbf{per-row local Top-$p$} selection rule adopted by nearly all training-free sparse attention methods.
Crucially, each head, and indeed each query row within a head, independently competes for its own budget.
The consequence of this isolation is the structural mismatch described in \S\ref{sec:motivation}: the same row-wise boundary is imposed on heads whose attention structures make that constraint very different in cost.

\subsection{Head Entropy as a Structural Descriptor}

The structural property that determines how a head responds to a larger competition domain is the \emph{concentration} of its attention distribution.
We quantify this through per-head attention entropy, computed directly from $P'$ at zero additional materialization cost:

\begin{equation}
    H_h = -\frac{1}{C_q}\sum_{i=1}^{C_q}\sum_{j=1}^{C_k} P'_{h,i,j} \cdot \log\!\big(P'_{h,i,j} + \varepsilon\big),
    \label{eq:entropy}
\end{equation}
where $\varepsilon$ is a small constant for numerical stability.
$H_h$ is the average per-row entropy: it is low when the attention mass within each query row concentrates on very few key clusters, and high when the mass spreads nearly uniformly across all key clusters.
For comparisons across configurations, we use the normalized entropy $\widetilde H_h=H_h/\log C_k$; normalization does not affect the within-layer ordering used by EG-Sparse.

\textbf{Why entropy is useful for domain assignment.}
Entropy describes the shape of the score landscape before sparse selection.
A low-entropy head concentrates importance on a small set of query--key cluster pairs; a high-entropy head distributes importance broadly.
As we demonstrate in the controlled diagnostic below, this difference determines whether a head benefits from or is harmed by a larger competition domain.
Entropy is therefore used not as a semantic head label, but as an online routing signal for domain assignment.

\subsection{The Competition Domain Mismatch}
\label{sec:motivation}

We isolate the competition-domain effect by holding the ranking score fixed and varying only the domain.
Let $\mathbf{V}_h\in\mathbb{R}^{C_k\times D}$ collect the Value centroid of each key cluster.
The dense centroid output of head $h$ is
\begin{equation}
    \mathbf{O}_h = \mathbf{P}'_h\mathbf{V}_h.
    \label{eq:dense_centroid_output}
\end{equation}
For a binary mask $\mathbf{M}\in\{0,1\}^{C_q\times C_k}$, we mask and renormalize each query row:
\begin{equation}
    \begin{aligned}
        \widetilde P'_{h,i,j}(\mathbf{M})
        &=
        \frac{P'_{h,i,j}M_{i,j}}
        {\sum_{\ell=1}^{C_k}P'_{h,i,\ell}M_{i,\ell}},\\
        \mathbf{O}_h(\mathbf{M})
        &=\widetilde{\mathbf{P}}'_h(\mathbf{M})\mathbf{V}_h .
    \end{aligned}
    \label{eq:sparse_centroid_output}
\end{equation}
Every mask considered below retains at least one key cluster per row, so the denominator is nonzero.
We measure centroid-output distortion using relative Frobenius error,
\begin{equation}
    e_h(\mathbf{M})
    =
    \frac{\lVert\mathbf{O}_h-\mathbf{O}_h(\mathbf{M})\rVert_F}
    {\lVert\mathbf{O}_h\rVert_F}.
    \end{equation}

\textbf{Domain hierarchy.}
We test four successively larger competition domains, all under the same Top-$p$ threshold, centroid construction, and $P'$ ranking score.
Only the set of candidates that compete for the same budget is varied:
\begin{enumerate}
    \item \textbf{Per-row local.} Each query row $(h,i)$ independently selects the top-$p$ portion of $P'_{h,i,:}$ as defined in Eq.~\eqref{eq:headwise}. This is the standard baseline adopted by existing training-free sparse methods and serves as the reference point for all comparisons.
    \item \textbf{Per-head global.} Within each head $h$, all $(i,j)$ pairs compete jointly by descending $P'_{h,i,j}$. This removes row boundaries but keeps each head independent.
    \item \textbf{Grouped cross-head.} Low-entropy heads compete in one shared pool; high-entropy heads compete in another. Heads are split by the median of their entropy $H_h$ (Eq.~\ref{eq:entropy}), using $P'$ as the ranking score within each pool.
    \item \textbf{All-head global.} All $H$ heads compete in a single pool over tuples $(h,i,j)$, ranked uniformly by $P'_{h,i,j}$.
\end{enumerate}
For the three expanded domains, which allocate budget globally rather than per row, we apply a minimum one-key-per-row safeguard: after Top-$p$ selection, any query row left empty receives its single highest-scoring key cluster. This affects fewer than $0.1\%$ of rows and ensures the renormalization in Eq.~\eqref{eq:sparse_centroid_output} always has a nonzero denominator.

\textbf{Error-change metric.}
Let $\mathbf{M}^{\mathrm{local}}_h$ denote the per-row local mask with error $e_h^{\mathrm{local}}$.
For each expanded domain, we compute a mask $\mathbf{M}^{\mathrm{domain}}_h$ at the same $p$ and measure the relative error change against the local baseline:
\begin{equation}
    \Delta_h^{\mathrm{local}}
    =
    \frac{e_h(\mathbf{M}^{\mathrm{domain}}) - e_h(\mathbf{M}^{\mathrm{local}})}
         {e_h(\mathbf{M}^{\mathrm{local}})}.
    \label{eq:delta}
\end{equation}
A negative $\Delta_h^{\mathrm{local}}$ means the expanded domain reduces reconstruction error relative to per-row local selection.

\begin{figure}[htbp]
\centering
\includegraphics[width=\columnwidth]{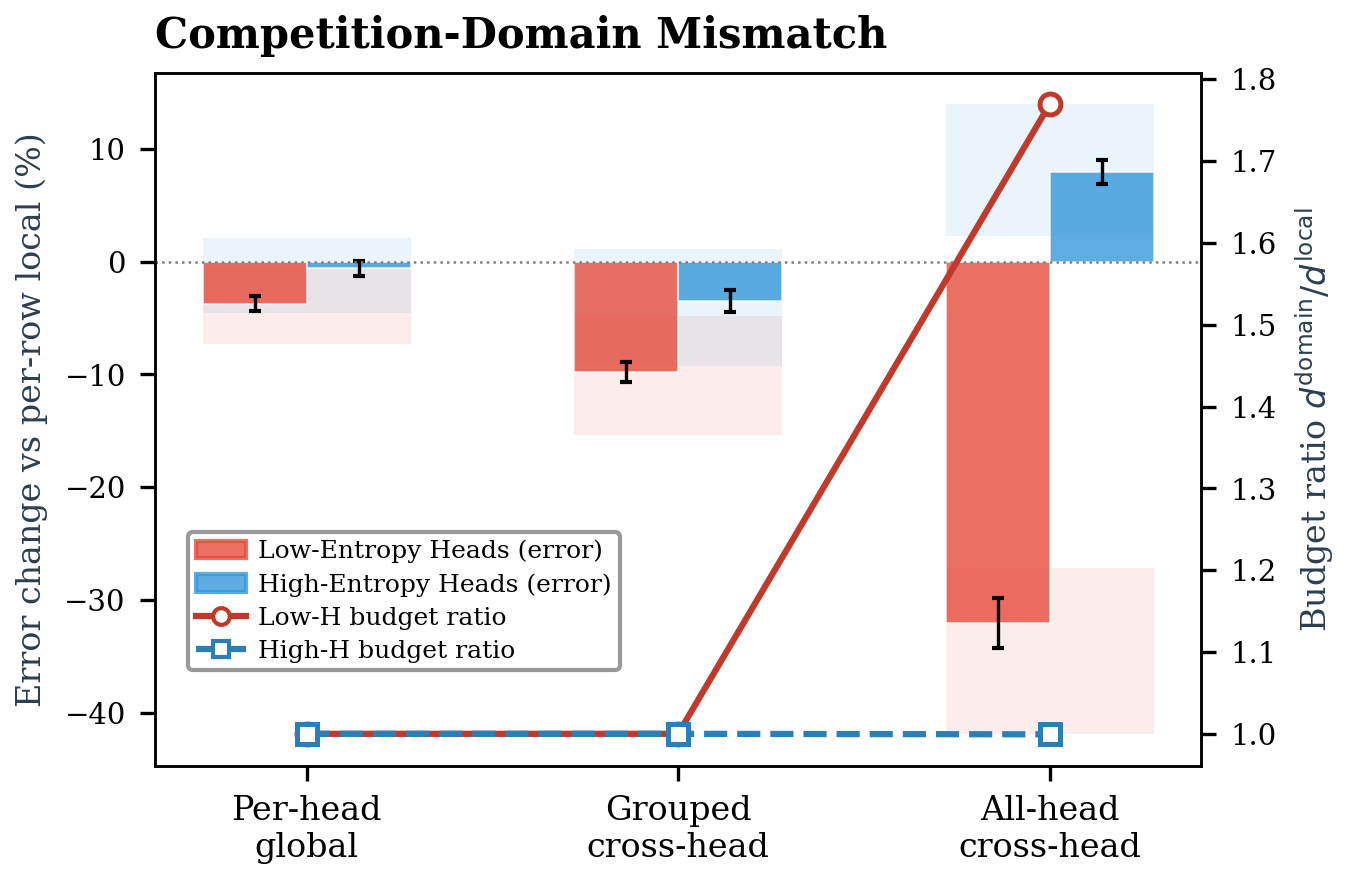}
\caption{\textbf{Competition-Domain Mismatch under a Fixed Ranking Score.} Bars report the relative reconstruction-error change from the per-row local baseline for low- and high-entropy heads; error bars show mean $\pm$ 95\% CI over 200 HunyuanVideo-T2V-13B prompts at 480P. The overlaid curves report the retained-budget ratio relative to the per-row local baseline. Low-entropy heads benefit increasingly from domain expansion, whereas high-entropy heads incur a $+7.9\%$ error increase under all-head global selection, where low-entropy heads capture a disproportionate share of the pooled budget.}
\label{fig:mismatch}
\end{figure}

\textbf{Results.}
The bars in Figure~\ref{fig:mismatch} report $\Delta_h^{\mathrm{local}}$ for the three expanded domains, split by low-entropy and high-entropy heads (median split on $H_h$).
We use $p{=}0.9$ and report measurements across 200 prompts from vbench on HunyuanVideo-T2V-13B at 480P, aggregated over six denoising steps and ten layers. Two patterns emerge.

First, at fixed scoring $P'$, changing only the domain alters reconstruction error by tens of percentage points, showing that the domain is an independent efficiency lever.

Second, the effect is head-dependent and asymmetric:
\begin{itemize}
    \item \textbf{Low-entropy heads benefit monotonically from domain expansion.}
          Per-head global reduces error by $3.7\%$; grouped cross-head reduces it further by $9.8\%$; all-head global achieves a $32.0\%$ reduction.
          Each successive domain enlargement improves reconstruction fidelity, indicating that per-row boundaries constrain these heads' efficiency.
    \item \textbf{High-entropy heads are indifferent to homogeneous expansion but are penalized under all-head global selection.}
          Per-head global and grouped cross-head produce changes near zero (${\approx}0\%$), while all-head global increases error by $+7.9\%$.
          When high-entropy and low-entropy heads compete in the same pool, the concentrated pairs from low-entropy heads capture a disproportionate share of the pooled budget, as shown by the overlaid budget-ratio curves in Figure~\ref{fig:mismatch}, crowding out high-entropy heads.
\end{itemize}

Fixed domains create opposing mismatches: per-row local selection prevents useful cross-row and cross-head redistribution for low-entropy heads, whereas all-head global selection causes interference for high-entropy heads.
We therefore use entropy to route only low-entropy heads to a grouped cross-head pool, retaining per-row local selection for the rest.

\subsection{Entropy-Guided Selective Domain Assignment}

EG-Sparse resolves this mismatch by making the competition domain an explicit, per-head, per-step design variable.
At each denoising step, the procedure operates in three stages.

\subsubsection{Stage 1: Automatic Thresholding}

The routing threshold $\tau$ is set to the median entropy across all $H$ heads:
\begin{equation}
    \tau = \text{median}\bigl(H_1, H_2, \ldots, H_H\bigr).
    \label{eq:median}
\end{equation}

The median requires \emph{zero manual tuning}, adapts to each layer and step's entropy scale, and is robust to outliers.
Table~\ref{tab:routing_ratio} confirms it as a strong calibration-free operating point.

\subsubsection{Stage 2: Head Splitting}

Heads are partitioned by their entropy relative to $\tau$:
\begin{align}
    \mathcal{H}^{\text{con}} &= \{\,h \mid H_h \leq \tau \,\}, \label{eq:con}\\
    \mathcal{H}^{\text{dif}} &= \{\,h \mid H_h > \tau \,\}. \label{eq:dif}
\end{align}
$\mathcal{H}^{\text{con}}$ contains heads with concentrated attention distributions; $\mathcal{H}^{\text{dif}}$ contains heads with diffuse attention.
Each group is routed according to its predicted domain benefit, as established by the diagnostic in Section~\ref{sec:motivation}.

\subsubsection{Stage 3: Domain-Specific Sparse Selection}

\textbf{Domain A: Grouped Cross-Head Top-$p$.}
For heads in $\mathcal{H}^{\text{con}}$, we construct a single competition pool over the joint index space:
\begin{equation}
    \Omega^{\text{con}} = \{\,(h,i,j) \mid h \in \mathcal{H}^{\text{con}},\; i =1,\dots,C_q,\; j =1,\dots,C_k\,\}.
\end{equation}
All tuples are ranked by the Value-aware importance score $U_{h,i,j} = P'_{h,i,j}\cdot\lVert \bm{c}^V_{h,j}\rVert_2$ in descending order, where $\bm{c}^V_{h,j}$ is the Value centroid of key cluster $j$ in head $h$. This score estimates the output contribution of each interaction: $U_{h,i,j}$ measures the L2 contribution of pair $(h,i,j)$ to the centroid output (Eq.~\ref{eq:dense_centroid_output}), and ranking by $U$ across heads directly prioritizes the tuples with the largest cumulative impact on total output fidelity.
Let $\pi$ be the descending permutation of $\Omega^{\text{con}}$ by $U$, and let $k^*$ be the smallest integer satisfying $\sum_{t=1}^{k^*} U_{\pi(t)} \ge p\!\!\sum_{\Omega^{\text{con}}}\!U$.
The retained tuples are the first $k^*$ elements of this sorted sequence:
\begin{equation}
    \mathcal{S}^{\text{con}} = \{\,\pi(1), \pi(2), \dots, \pi(k^*)\,\},
    \label{eq:global_select}
\end{equation}
A binary mask $\mathbf{M}^{\text{con}}_h \in \{0,1\}^{C_q \times C_k}$ is set for each $h \in \mathcal{H}^{\text{con}}$.

This shared domain replaces many independent stopping rules with one joint allocation over candidate tuples.
It does not reuse an interaction across rows; rather, it allows the retained-pair budget to move toward the highest-importance $(h,i,j)$ tuples without requiring every row and head to satisfy an identical threshold independently.
We additionally enforce a minimum one-key-per-row constraint: after Top-$p$ selection, any query row left empty is filled with its single highest-scoring key cluster.
In practice, this fallback affects fewer than $0.1\%$ of rows and does not change density at the reported precision.

\textbf{Domain B: Per-Row Local Top-$p$.}
Heads in $\mathcal{H}^{\text{dif}}$ retain the standard per-row local selection of Eq.~\eqref{eq:headwise}.
Because high-entropy heads show no benefit from domain expansion in the diagnostic (Section~\ref{sec:motivation}), coupling them to the grouped cross-head pool adds complexity without a comparable expected efficiency gain.

\subsubsection{Mask Merging and Sparse Execution}

The two masks are merged into a single per-head mask $\mathbf{M} \in \{0,1\}^{H \times C_q \times C_k}$:
\begin{equation}
    \mathbf{M}_h = \mathbf{M}^{\text{con}}_h \;\text{for}\; h \in \mathcal{H}^{\text{con}},\qquad
    \mathbf{M}_h = \mathbf{M}^{\text{dif}}_h \;\text{for}\; h \in \mathcal{H}^{\text{dif}}.
\end{equation}
The merged mask is expanded to token granularity using the cluster sizes and handed to an unmodified variable block-sparse attention kernel.
The kernel operates identically regardless of which domain produced each head's mask.
EG-Sparse introduces zero parameters, zero calibration data, and zero offline preprocessing.

\subsection{Value-Aware Ranking in the Cross-Head Pool}

The competition-domain diagnostic in Section~\ref{sec:motivation} uses attention mass $P'$ to isolate the domain effect.
In the deployed method, the cross-head pool replaces $P'$ with a value-aware importance score:
\begin{equation}
    U_{h,i,j} = P'_{h,i,j} \cdot \lVert \bm{c}^V_{h,j}\rVert_2,
    \label{eq:u_score}
\end{equation}
where $\bm{c}^V_{h,j}$ is the value centroid of key cluster $j$ in head $h$.
This score estimates each pair's contribution to the centroid output by scaling attention mass by the value-vector norm.
Pairs with large attention mass but small output impact are deprioritized relative to pairs whose selection moves the output substantially.
The domain routing (which heads share a pool) and the ranking score (how pairs are sorted within that pool) are separate design choices; our ablation in Table~\ref{tab:routing} confirms that domain routing alone provides a measurable gain (PSNR $21.79$ vs.\ $21.56$ for per-row local), and value-aware ranking amplifies this within the already-selected domain (PSNR $21.95$).

\section{Experiments}
\begin{table*}[htbp]
\centering
\setlength{\tabcolsep}{4pt}
\caption{\textbf{Main results in the evaluated model configurations.} All sparse methods use no dense warmup. Density is the fraction of retained centroid pairs; lower is better. GPU denoising latency is the total GPU time for the 50-step DiT denoising process, and speedup is relative to Dense. $\dagger$: $p<0.05$ vs.\ SVG2 (Wilcoxon).}
\label{tab:main}
\small
\begin{tabular}{llccccccc}
\toprule
\textbf{Model} & \textbf{Method} & \textbf{PSNR}$\uparrow$ & \textbf{SSIM}$\uparrow$ & \textbf{LPIPS}$\downarrow$ & \textbf{VBench}$\uparrow$ & \textbf{Density}$\downarrow$ & \textbf{Latency (s)}$\downarrow$ & \textbf{Speedup vs.\ Dense}$\uparrow$ \\
\midrule
\multirow{5}{*}{\shortstack[l]{\texttt{Wan2.1}\\\texttt{I2V-14B}\\\texttt{720P}}}
 & Dense          & --   & --   & --   & 79.31 & 100\% & 1823 & 1.00$\times$ \\
 & SVG            & 17.80 & 0.635 & 0.318 & 77.40 & 28\%  & 974  & 1.87$\times$ \\
 & SpargeAttn     & 17.60 & 0.628 & 0.325 & 77.20 & 41\%  & 1050 & 1.74$\times$ \\
 & SVG2           & 18.26 & 0.649 & 0.307 & 77.73 & 27\%  & 902  & 2.02$\times$ \\
 & \textbf{EG-Sparse (Ours)}  & \textbf{18.75}$^{\dagger}$ & \textbf{0.672} & \textbf{0.252} & \textbf{78.52} & \textbf{25\%} & \textbf{889} & \textbf{2.05$\times$} \\
\midrule
\multirow{6}{*}{\shortstack[l]{\texttt{Wan2.1}\\\texttt{T2V-14B}\\\texttt{720P}}}
 & Dense          & --   & --   & --   & 78.16 & 100\% & 1815 & 1.00$\times$ \\
 & SVG            & 14.96 & 0.575 & 0.472 & 74.45 & 30\%  & 975  & 1.86$\times$ \\
 & SpargeAttn     & 13.30 & 0.314 & 0.706 & 74.21 & 42\%  & 1045 & 1.74$\times$ \\
 & XAttention     & 16.01 & 0.623 & 0.505 & 76.00 & 29\%  & 893  & 2.03$\times$ \\
 & SVG2           & 16.47 & 0.611 & 0.467 & 76.03 & 29\%  & 904  & 2.01$\times$ \\
 & \textbf{EG-Sparse (Ours)}  & \textbf{17.03}$^{\dagger}$ & \textbf{0.655} & \textbf{0.446} & \textbf{76.24} & \textbf{27\%} & \textbf{881} & \textbf{2.06$\times$} \\
\midrule
\multirow{5}{*}{\shortstack[l]{\texttt{HunyuanVideo}\\\texttt{T2V-13B}\\\texttt{720P}}}
 & Dense          & --   & --   & --   & 78.50 & 100\% & 1620 & 1.00$\times$ \\
 & SVG            & 12.21 & 0.483 & 0.491 & 75.89 & 28\%  & 646  & 2.51$\times$ \\
 & XAttention     & 17.25 & 0.637 & 0.421 & 76.07 & 29\%  & 689  & 2.35$\times$ \\
 & SVG2           & 17.39 & 0.648 & 0.414 & 76.12 & 28\%  & 611  & 2.65$\times$ \\
 & \textbf{EG-Sparse (Ours)}  & \textbf{17.93}$^{\dagger}$ & \textbf{0.716} & \textbf{0.394} & \textbf{76.36} & \textbf{26\%} & \textbf{602} & \textbf{2.69$\times$} \\
\midrule
\multirow{6}{*}{\shortstack[l]{\texttt{HunyuanVideo}\\\texttt{T2V-13B}\\\texttt{480P}}}
 & Dense          & --   & --   & --   & 80.10 & 100\% & 415  & 1.00$\times$ \\
 & SVG            & 18.64 & 0.673 & 0.291 & 77.89 & 28\%  & 237  & 1.75$\times$ \\
 & SpargeAttn     & 18.70 & 0.665 & 0.310 & 77.94 & 39\%  & 254  & 1.63$\times$ \\
 & XAttention     & 19.20 & 0.690 & 0.290 & 78.01 & 29\%  & 243  & 1.71$\times$ \\
 & SVG2           & 19.88 & 0.700 & 0.281 & 78.43 & 27\%  & 218  & 1.90$\times$ \\
 & \textbf{EG-Sparse (Ours)}  & \textbf{20.35}$^{\dagger}$ & \textbf{0.740} & \textbf{0.252} & \textbf{78.62} & \textbf{24\%} & \textbf{204} & \textbf{2.03$\times$} \\
\bottomrule
\end{tabular}
\end{table*}
Our evaluation is organized around three questions: (i)~Does EG-Sparse improve quality--efficiency across the evaluated video-DiT configurations (\S\ref{sec:main})? (ii)~Does the competition-domain mismatch diagnosed in Section~\ref{sec:motivation} translate to practical gains, and is selective domain expansion necessary (\S\ref{sec:mechanism})? (iii)~What overhead does online domain adaptation introduce (\S\ref{sec:overhead})?

\subsection{Experimental Setup}
\label{sec:setup}

\paragraph{Models and data.}
We evaluate four model configurations spanning two model families, text- and image-conditioned generation, and two resolutions:
Wan2.1-I2V-14B at 720P,
Wan2.1-T2V-14B at 720P,
HunyuanVideo-T2V-13B at 480P,
HunyuanVideo-T2V-13B at 720P.
We use Wan2.1-T2V-1.3B at 480P for the selective-policy ablation and HunyuanVideo-T2V-13B at 480P for the domain diagnostic and routing-ratio sensitivity.
Following standard protocol~\cite{svg2_2025}, HunyuanVideo-T2V-13B samples are generated with 129 frames, Wan2.1-T2V-14B samples with 81 frames, all using 50 denoising steps.
Sparse attention is applied at every denoising step with no dense warmup (\texttt{warmup}=0).
Evaluation uses the standard VBench benchmark~\cite{vbench} across all configurations.
Experiments are conducted on NVIDIA GH200 GPUs.

\paragraph{Metrics.}
We report three orthogonal families:
(a) \textbf{fidelity to dense attention}: PSNR, SSIM, LPIPS computed frame-by-frame against the dense reference output;
(b) \textbf{perceptual quality}: VBench~\cite{vbench} overall score, following the official weighting protocol;
(c) \textbf{efficiency}: attention density, the fraction of the full attention matrix computed on the GPU; GPU denoising latency for the complete 50-step DiT denoising process; and speedup over dense inference.

\paragraph{Baselines.}
We compare against \textbf{Dense} (full 3D attention), \textbf{SVG}~\cite{svgen}, \textbf{SpargeAttn}~\cite{spargeattn2025}, \textbf{SVG2}~\cite{svg2_2025} (the base framework we build on), and \textbf{XAttention}~\cite{xattention2025}.
For the mechanism ablation, we additionally construct \textbf{All-head global} (all 12 heads compete in a single Top-$p$ pool), \textbf{Random Split} (each step randomly assigns half the heads to grouped cross-head selection and the rest to per-row local selection), and \textbf{Fixed Split} (heads 0--5 permanently use grouped cross-head selection, while heads 6--11 use per-row local selection).

\paragraph{Configuration.}
Unless stated otherwise, we use Top-$p$ threshold $p{=}0.9$ for all methods.
The grouped cross-head competition domain uses Value-aware Importance $U_{h,i,j} = P'_{h,i,j}\cdot\lVert \bm{c}^V_{h,j}\rVert_2$ as the ranking score.

\paragraph{Competition-domain diagnostic.}
Figure~\ref{fig:mismatch} is computed on 200 T2V prompts with HunyuanVideo-T2V-13B at 480P.
We sample six denoising steps $\{0,9,18,27,36,45\}$ and ten layers $\{0,6,12,18,24,30,36,42,48,54\}$, with 24 heads per layer.
For each sampled step--layer--head tuple, we compute the relative error change $\Delta_h^{\mathrm{local}}$ (Eq.~\ref{eq:delta}) for the three expanded domains against the per-row local baseline.
Reported values aggregate all sampled steps and layers, with heads grouped into low-entropy and high-entropy halves by their entropy $H_h$.

\paragraph{Statistical rigor.}
We apply a paired Wilcoxon signed-rank test on per-prompt PSNR against SVG2; $\dagger$ denotes $p < 0.05$.
For VBench, we follow the official evaluation protocol with the full benchmark prompt set and report the weighted overall score.

\subsection{Main Results: Fidelity and Efficiency}
\label{sec:main}

Table~\ref{tab:main} shows EG-Sparse improves fidelity and efficiency across model families and conditioning modes. Across all four configurations, EG-Sparse achieves the best PSNR, SSIM, LPIPS, and VBench among the compared sparse methods while also attaining the lowest attention density and GPU denoising latency, with $2.03\times$--$2.69\times$ speedup over Dense.
This lower-density, higher-quality result shows that the gain does not come from retaining more interactions; instead, domain adaptation allocates the sparse-attention budget more effectively across rows and heads according to their structural needs.
On HunyuanVideo, the PSNR and VBench margins over the strongest competing sparse baseline are larger at 720P than at 480P, suggesting that the benefit may become more pronounced as spatial resolution and attention demands increase.
The following mechanism analysis further isolates the contributions of selective domain routing and value-aware ranking.

\subsection{Mechanism Analysis: The Competition-Domain Effect}
\label{sec:mechanism}

The controlled diagnostic in Section~3.3 established that the competition-domain effect is head-dependent and asymmetric: low-entropy heads benefit from domain expansion while high-entropy heads are penalized under all-head global selection. Table~\ref{tab:routing} tests whether selective, entropy-guided domain adaptation converts this diagnosis into practical quality gains over uniform and non-adaptive domain policies.

\begin{table}[htbp]
\centering
\setlength{\tabcolsep}{4pt}
\caption{\textbf{Selective-globalization ablation on Wan2.1-T2V-1.3B at 480P.} All variants share the same upstream candidate representation and use $p{=}0.9$. $\dagger$: $p<0.05$ vs.\ SVG2 (Wilcoxon).}
\label{tab:routing}
\small
\begin{tabular}{lccc}
\toprule
\textbf{Selection Rule} & \textbf{PSNR}$\uparrow$ & \textbf{VBench}$\uparrow$ & \textbf{Speedup}$\uparrow$ \\
\midrule
SVG2: per-row local       & 21.56 & 76.79 & 1.15$\times$ \\
All-head global           & 21.47 & 76.69 & 1.18$\times$ \\
Random 50/50 split        & 21.18 & 76.41 & 1.16$\times$ \\
Fixed head-index split    & 21.28 & 76.52 & 1.16$\times$ \\
\textbf{Entropy split}    & \textbf{21.95}$^{\dagger}$ & \textbf{77.35} & 1.13$\times$ \\
\ \ \ \ \ w/o Value weighting & 21.79$^{\dagger}$ & 77.15 & 1.14$\times$ \\
\bottomrule
\end{tabular}
\end{table}

Table~\ref{tab:routing} supports three conclusions.
First, all-head global underperforms per-row local (PSNR $21.47$ vs.\ $21.56$), consistent with the observed $+7.9\%$ high-entropy penalty under all-head global selection; unselective expansion is harmful.
Second, the entropy split without value weighting (PSNR $21.79$ vs.\ $21.56$, $+0.23$\,dB) significantly improves over the per-row local baseline, confirming that domain routing alone provides a measurable gain.
Because both the per-row local domain and the grouped cross-head pool in this variant use the same $P'$ ranking score, this gain is attributable to the domain assignment rather than to an improved scoring function.
Third, adding value-aware scoring in the cross-head pool further raises PSNR to $21.95$.
The additional $0.16$\,dB gain from $21.79$ to $21.95$ represents the benefit of ranking cross-head pairs by their output contribution rather than by raw attention mass.
Random and fixed splits both underperform the per-row local baseline (PSNR $21.18$ and $21.28$), confirming that entropy is the operative domain-selection signal rather than any arbitrary head split.
\begin{table}[htbp]
\centering
\setlength{\tabcolsep}{5pt}
\caption{\textbf{Sensitivity to the fraction of low-entropy heads routed to grouped cross-head selection} on HunyuanVideo\-T2V\-13B at 480P with $p{=}0.9$. Heads are added in ascending entropy order. Fidelity improves as heads predicted to benefit from grouped cross-head selection enter the pool, peaks at the median split, and declines as the pool expands into the higher-entropy regime where inter-group budget interference outweighs the gain.}
\label{tab:routing_ratio}
\small
\begin{tabular}{lccc}
\toprule
\textbf{Grouped Cross-Head (of 24)} & \textbf{PSNR}$\uparrow$ & \textbf{SSIM}$\uparrow$ & \textbf{LPIPS}$\downarrow$ \\
\midrule
6$\;$($25\%$) & 19.86 & 0.701 & 0.280 \\
10 ($40\%$) & 20.04 & 0.700 & 0.272 \\
\textbf{12 ($50\%$)} & \textbf{20.35} & \textbf{0.740} & \textbf{0.252} \\
14 ($60\%$) & 20.11 & 0.710 & 0.267 \\
18 ($75\%$) & 19.78 & 0.708 & 0.277 \\
\bottomrule
\end{tabular}
\end{table}
\paragraph{Sensitivity to the routing ratio.}
The median threshold routes approximately half of the heads with the lowest entropy to the grouped cross-head domain.
To test whether this choice requires fine tuning, we vary the number of low-entropy heads assigned to the cross-head pool (HunyuanVideo-T2V-13B has 24 heads per layer) while keeping $p{=}0.9$.
Table~\ref{tab:routing_ratio} shows that the median split (12 heads) achieves the best PSNR, SSIM, and LPIPS simultaneously.
Moving from 6 to 12 heads progressively captures more heads that benefit from grouped cross-head selection, but extending the pool to 14 and 18 heads introduces increasingly higher-entropy heads and reverses the fidelity gain.
This saturation pattern matches the diagnostic in Figure~\ref{fig:mismatch}: the benefit of domain expansion is concentrated in low-entropy heads, and including higher-entropy heads eventually triggers the cross-group budget interference indicated by the overlaid budget-ratio curves.
The median is therefore a strong calibration-free operating point in the evaluated setting rather than an arbitrary routing ratio.

\subsection{Efficiency and Overhead Analysis}
\label{sec:overhead}

\begin{table}[htbp]
\centering
\setlength{\tabcolsep}{2pt}
\caption{\textbf{Per-attention-block latency breakdown on HunyuanVideo-T2V-13B at 480P (GH200, ms).} Computing the entropy signal costs only $0.13$\,ms; the complete selective domain routing and sparse execution costs are reported separately.}
\label{tab:latency}
\small
\begin{tabular}{p{0.52\columnwidth}rr}
\toprule
\textbf{Component} & \textbf{Latency (ms)} & \textbf{Share} \\
\midrule
Candidate preprocessing
    & 13.64 & 41.8\% \\
Head-entropy computation
    & 0.13 & 0.4\% \\
Selective domain routing and mask selection
    & 2.80 & 8.6\% \\
FlashInfer sparse attention
    & 16.06 & 49.2\% \\
\midrule
\textbf{Total per attention block}
    & \textbf{32.63} & \textbf{100\%} \\
\bottomrule
\end{tabular}
\end{table}

The entropy signal and selective domain routing account for $9.0\%$ of per-block attention latency (Table~\ref{tab:latency}); Table~\ref{tab:main} includes these costs.
The breakdown also explains why density does not translate one-for-one into denoising speed: shared candidate preprocessing consumes $41.8\%$ of attention-block time, and computation outside sparse attention is unchanged.
Even with these unchanged costs, EG-Sparse reduces GPU denoising latency by $1.4\%$--$6.4\%$ relative to SVG2 across the four evaluated configurations.

\section{Conclusion}

We revisited a design choice shared by training-free sparse attention methods for video diffusion transformers: fixing sparse competition independently within every query row.
A controlled diagnostic at fixed ranking score shows that the cost of this uniform boundary is head-dependent and asymmetric: low-entropy heads benefit monotonically from larger competition domains, while high-entropy heads are penalized when forced into the same pool.

EG-Sparse adapts domains online, using current head entropy at every layer and denoising step to route low-entropy heads to a grouped cross-head pool while retaining per-row local selection for the rest.
The median rule requires neither retraining, offline calibration, nor manual threshold selection.
Across the evaluated bidirectional video DiTs, this selective expansion reduces attention density and GPU denoising latency while improving overall dense-reference fidelity and generation quality; its behavior in causal or non-diffusion Transformers is untested.
Ablations show that the gain is not explained by uniform expansion or an arbitrary split: it depends on directing expansion toward heads predicted to benefit.

The broader implication is that video-DiT sparse attention should optimize not only the score and budget, but also the domain over which candidates compete.
Together, these findings support competition-domain adaptation as a useful design principle for training-free sparse attention in video DiTs.

\bibliography{aaai2027}

\end{document}